\documentclass[conference]{IEEEtran}
\IEEEoverridecommandlockouts
\usepackage{cite}
\usepackage{amsmath,amssymb,amsfonts}
\usepackage{algorithmic}
\usepackage{graphicx}
\usepackage{textcomp}
\usepackage{xcolor}
\usepackage{algorithm}
\usepackage{algorithmic}
\usepackage{url}
\usepackage{hyperref}

\usepackage{etoolbox}
\makeatletter
\patchcmd{\@makecaption}
  {\scshape}
  {}
  {}
  {}
\makeatother

\def\BibTeX{{\rm B\kern-.05em{\sc i\kern-.025em b}\kern-.08em
    T\kern-.1667em\lower.7ex\hbox{E}\kern-.125emX}}

\makeatletter
\newcommand{\linebreakand}{%
  \end{@IEEEauthorhalign}
  \hfill\mbox{}\par
  \mbox{}\hfill\begin{@IEEEauthorhalign}
}
\makeatother

\IEEEoverridecommandlockouts
\begin{document}

\title{
%No Need to be Subtle: 
Attacking Vision-based Perception in End-to-End Autonomous Driving Models
\thanks{This research was partially supported by NSF grants CNS-1739643, IIS-1905558 and CNS-1640624, ARO grant W911NF1610069 and MURI grant W911NF1810208.}}

\author{
    \IEEEauthorblockN{
    Adith Boloor\IEEEauthorrefmark{1}, 
    Karthik Garimella\IEEEauthorrefmark{1}, 
    Xin He\IEEEauthorrefmark{2},
    Christopher Gill\IEEEauthorrefmark{1}, 
    Yevgeniy Vorobeychik\IEEEauthorrefmark{1}, 
    Xuan Zhang\IEEEauthorrefmark{1}}
    \IEEEauthorblockA{\IEEEauthorrefmark{1}Washington University in St. Louis
    \\\{adith, kvgarimella, cdgill, yvorobeychik, xuan.zhang\}@wustl.edu}
    \IEEEauthorblockA{\IEEEauthorrefmark{2}University of Michigan, Ann Arbor
    \\\{xinhe\}@umich.edu}
}
% \author{\IEEEauthorblockN{1\textsuperscript{st} Given Last}
% \author{\IEEEauthorblockN{Adith Boloor\IEEEauthorrefmark{1}}
% \IEEEauthorblockA{\textit{Electrical and Systems Engineering}\\
% \textit{Washington University in St. Louis}\\
% St. Louis, USA \\
% a.jagadi@wustl.edu}

% \and

% \IEEEauthorblockN{Karthik Garimella\IEEEauthorrefmark{1}}
% \IEEEauthorblockA{\textit{Computer Science and Engineering} \\
% \textit{Washington University in St. Louis}\\
% St. Louis, USA \\
% kvgarimella@wustl.edu}

% \and

% \IEEEauthorblockN{Xin He\IEEEauthorrefmark{2}}
% \IEEEauthorblockA{\textit{Computer Science and Engineering} \\
% \textit{University of Michigan, Ann Arbor}\\
% Ann Arbor, USA \\
% xinhe@umich.edu}
% \linebreakand % check \newcommand before begin{document}

% \IEEEauthorblockN{Christopher Gill\IEEEauthorrefmark{1}}
% \IEEEauthorblockA{\textit{Computer Science and Engineering} \\
% \textit{Washington University in St. Louis}\\
% St. Louis, USA \\
% cdgill@wustl.edu}

% \and

% \IEEEauthorblockN{Yevgeniy Vorobeychik\IEEEauthorrefmark{1}}
% \IEEEauthorblockA{\textit{Computer Science and Engineering} \\
% \textit{Washington University in St. Louis}\\
% St. Louis, USA \\
% yvorobeychik@wustl.edu}

% \and

% \IEEEauthorblockN{Xuan Zhang\IEEEauthorrefmark{1}}
% \IEEEauthorblockA{\textit{Electrical and Systems Engineering} \\
% \textit{Washington University in St. Louis}\\
% St. Louis, USA \\
% xuan.zhang@wustl.edu}
% }

\maketitle

\begin{abstract}
Recent advances in machine learning, especially techniques such as deep neural networks, are enabling a range of emerging applications. One such example is autonomous driving, which often relies on deep learning for perception.
However, deep learning-based perception has been shown to be vulnerable to a host of subtle adversarial manipulations of images.
Nevertheless, the vast majority of such demonstrations focus on perception that is disembodied from end-to-end control.
We present novel end-to-end attacks on autonomous driving in simulation, using simple physically realizable attacks: the painting of black lines on the road.
These attacks target deep neural network models for end-to-end autonomous driving control.
A systematic investigation shows that such attacks are easy to engineer, and we describe scenarios (e.g., right turns) in which they are highly effective. We define several objective functions that quantify the success of an attack and develop techniques based on Bayesian Optimization to efficiently traverse the search space of higher dimensional attacks. 
Additionally, we define a novel class of \emph{hijacking} attacks, where painted lines on the road cause the driverless car to follow a target path. 
Through the use of network deconvolution, we provide insights into the successful attacks, which appear to work by mimicking activations of entirely different scenarios. Our code is available on
\href{https://github.com/xz-group/AdverseDrive}{\textit{https://github.com/xz-group/AdverseDrive}}
\end{abstract}


\section{Introduction}

\begin{figure}
  \centering  \includegraphics[width=0.9\columnwidth]{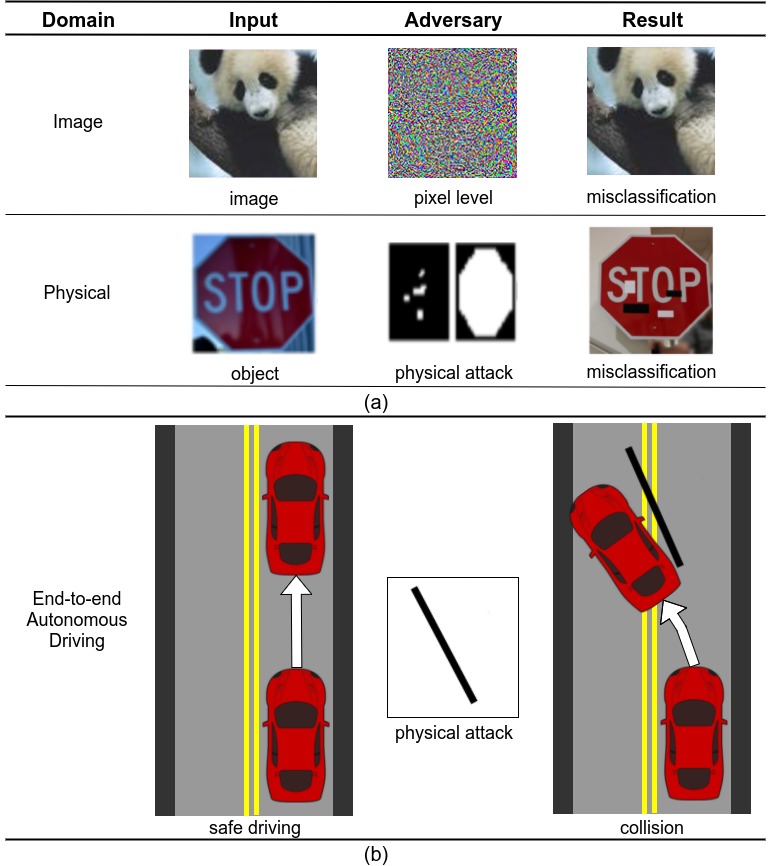}
  \caption{(a) Existing attacks on machine learning models in the image ~\cite{Goodfellow2015ExplainingAH} and the physical domain ~\cite{Eykholt2017RobustPA}; (b) conceptual illustration of potential physical attacks in the end-to-end driving domain studied in our work.}
  \label{fig:conceptual_overview}
\end{figure}

With billions of dollars being pumped into autonomous vehicle research to reach Level 5 Autonomy, where vehicles will not require human intervention, safety has become a critical issue \cite{1906.02939}.
Remarkable advances in deep learning, in turn, suggest such approaches as natural candidates for integration into autonomous control.
One way to use deep learning in autonomous driving control is in an end-to-end (e2e) fashion, where learned models directly translate perceptual inputs into control decisions, such as the vehicle's steering angle.
Indeed, recent work demonstrated such approaches to be remarkably successful, particularly when learned to imitate human drivers~\cite{Bojarski2017ExplainingHA}.
%, the most prevalent issue that researchers are focusing on is improving safety of the passengers, pedestrians and other commuters. Despite using abundant sensors and training over millions of miles, current autonomous capabilities in vehicles are still far from robust in diverse driving and environmental conditions. Advances in deep neural networks (DNNs) have allowed end-to-end (e2e) learning models to generate steering controls directly from sensory information \cite{Bojarski2017ExplainingHA}. Despite its superior capability of mimicking driving behaviors, DNNs remain fragile to unseen scenarios, as well as deliberate adversaries.

Despite the success of deep learning in enabling greater autonomy, a number of parallel efforts also have exhibited concerning fragility of deep learning approaches to small adversarial perturbations of inputs such as images ~\cite{Vorobeychik18book,Dreossi2018SemanticAD}.
Moreover, such perturbations have been shown to effectively translate to physically realizable attacks on deep models, such as placing stickers on stop signs to cause miscategorization of these as speed limit signs~\cite{Eykholt2017RobustPA}.
Fig.~\ref{fig:conceptual_overview}(a) offers several canonical illustrations.

There is, however, a crucial missing aspect of most adversarial example attacks to date: manipulations of the physical environment that have a demonstrable \emph{physical} impact (e.g., a crash).
For example, typical attacks consider only prediction error as an outcome measure and focus either on a static image, or a fixed set of views, without consideration of the dynamics of closed-loop autonomous control.
To bridge this gap, our aim is to study  \emph{end-to-end} adversarial examples.
We require such adversarial examples to: 1) modify the physical environment, 2) be simple to implement, 3) appear unsuspicious, and 4) have a physical impact, such as causing an infraction (lane violation or collision).
The existing attacks that introduce carefully engineered manipulations fail the simplicity criterion~\cite{Papernot2016TheLO,Vorobeychik18book}, whereas the simpler physical attacks, such as stickers on a stop sign, are evaluated solely on prediction accuracy~\cite{Eykholt2017RobustPA}.

The particular class of attacks we systematically study is the painting of black lines on the road, as shown in Fig.~\ref{fig:conceptual_overview}(b).
These are unsuspicious since they are semantically inconsequential (few human drivers would be confused) and are similar to common imperfections observed in the real world, such as skid marks or construction markers.
Furthermore, we demonstrate a systematic approach for designing such attacks so as to maximize a series of objective functions, and demonstrate actual physical impact (lane violations and crashes) over a variety of scenarios, in the context of end-to-end deep learning-based controllers in the CARLA autonomous driving simulator~\cite{Dosovitskiy2017CARLAAO}. 

We consider scenarios where correct behavior involves turning right, left, and driving straight.
Surprisingly, we find that right turns are by far the riskiest, meaning that the right scenario is the easiest to attack; on the other hand, as expected, going straight is comparatively robust to our class of attacks. 
We use network deconvolution to explore the reasons behind successful attacks.
Here, our findings suggest that one of the causes of controller failure is partially mistaking painted lines on the road for a curb or barrier common during left-turn scenarios, thereby causing the vehicle to steer sharply left when it would otherwise turn right.  By increasing the dimensionality of our attack space and using a more efficient Bayesian optimization strategy, we are able to find successful attacks even for cases where the driving agent needs to go straight. 
%Our final contribution examines intersection scenarios where attacks are able to cause the agent to take a turn designed by an attacker rather than the expected turn (e.g. take a left turn instead of a right turn), effectively hijacking the vehicle.  
Our final contribution is a demonstration of novel \emph{hijacking} attacks, where painting black lines on the road causes the car to follow a target path, even when it is quite different from the correct route (e.g., causing the car to turn left instead of right).

This paper is an extension our previous work \cite{simple-physical-adversaries}, with the key additions of new objective functions, a new optimization strategy, Bayesian Optimization, and a new type of adversary in the form of hijacking self-driving models.
In this paper, we first talk about relevant prior work on deep neural networks, adversarial machine learning in the context of autonomous vehicles, in Section \ref{sec:background_and_related_work}. Then in Section \ref{sec:modeling_framework} we define the problem statement and present several objective functions that mathematically represent the problem statement. In Section \ref{sec:approaches_for_generating_adversaries}, we introduce some optimization strategies. In Section \ref{sec:experimental_methodology}, we discuss our experimental setup including our adversary generation library and simulation pipeline. Section \ref{sec:experiment_results} shows how we were able to successfully generate adversaries against e2e models, and presents a new form of attack, dubbed the hijacking attack where we control the route of the e2e model. 
% Finally, we discuss our results, including how some attacks are more effective in some cases than others, in Section \ref{sec:discussion}.

\section{Related Work}
\label{sec:background_and_related_work}

%This section provides the foundations of the various concepts and frameworks used within this paper. 
%We begin by providing the background and related work.
%First, we introduce neural network models and in particular, deep neural networks for perception. We then discuss attacks on these models, and procedures for generating successful attacks using a probabilistic approach. 

\subsection{Deep Neural Networks for Perception and Control}
\label{subsec:dnn_perception_control}
Neural Networks (NN) are machine learning models that consist of multiple layers of neurons, where each neuron implements a simple function (such as a sigmoid function), and where the output is a prediction.
%are loosely modeled after the human brain which allows them to recognize patterns in high-dimensional data.  
%The simplest NN consists of an input layer that accepts input parameters, a hidden layer that interprets the previous layer using an activation function, and an output layer, which generates the output of the model. 
%To address large complex problems, 
Deep Neural Networks (DNNs) are neural networks with more than two layers of neurons, and have come to be the state-of-the-art approach for a host of problems in perception, such as image classification and semantic segmentation \cite{Krizhevsky2012ImageNetCW}. 
%are designed with a deeper and wider hierarchy (at least more than one hidden layer) so that the network model has a larger learning capability to accommodate diverse inputs with more features. 
%They have been used to achieve a high level of accuracy in perception related tasks such as 
Despite having complete autonomous driving stacks which include trained DNN models for perception, a series of real-world crashes involving autonomous vehicles demonstrate the stakes, and some of the existing limitations of the technology \cite{tesla_autopilot_crash, uber_selfdriving_crash}.
%, where self-driving vehicles crashed and even fatally wounded pedestrians. To ensure the safety of such systems, there is a strong need to improve the perception modules of self-driving vehicles.

% tasks like ImageNet challenges , more hidden layers, different kinds of layers (e.g., convolutional and batchnorm layers) and advanced techniques (e.g., Resnet\cite{He2016DeepRL}) are incorporated to extract different features from the input, explore their relations, and then converge to a desirable output. DNNs are being used extensively in the field of autonomous vehicles, but their limitations in this scenario have not been thoroughly studied. 

\subsection{End-to-end Deep Learning}

End-to-end (e2e) learning models are comprised of DNNs that accept raw input parameters in one end and directly calculate the desired output at the other end. Rather than explicitly decomposing a complex problem into its constituent parts and solving them separately, e2e models directly generate the output from the inputs. It is achieved by applying gradient-based learning to the system as a whole. Recently, e2e models have been shown to have good performance in the domain of autonomous vehicles, where the forward facing camera input can be directly translated to control (steer, throttle and brake) commands~\cite{Bojarski2016EndTE, DBLP:journals/corr/ZhangC16b, end_to_end_lane_keeping, Xu_2017_CVPR}. 
% Since these models still rely on DNNs, they are vulnerable to adversarial attacks \cite{Glasmachers2017LimitsOE}.

\subsection{Attacks on Deep Learning for Perception and Control}
\label{subsec:attacks_on_deep_learning}

Adversarial examples (also called attacks, and adversaries) \cite{Vorobeychik18book,1904.07370,1801.00553, Lowd2005AdversarialL} are deliberately calculated perturbations to the input which result in an error in the output from a trained DNN model. 
% NNs exhibits a superior accuracy when an input is similar to what it has been trained for, but with adversarial attacks, it is shown that NNs are not as robust as we believed them to be. 
% The state-of-the-art adversarial attack technique uses fast gradient descent methods to firstly figure out what parts of the input data are more sensitive to the perturbation and then intentionally adds a unnoticeable noise to those pixels to create an erroneous output. 
The idea of using adversarial examples against static image classification models demonstrated that DNNs are highly susceptible to carefully designed pixel-level adversarial perturbations\cite{Papernot2016TheLO, Goodfellow2014GenerativeAN,Vorobeychik18book}. 
%Perturbed images that would be easily ignored by humans may not be correctly recognized by the DNN model. 
More recently, adversarial example attacks have been implemented in the physical domain~\cite{Eykholt2017RobustPA, Lu2017NONT, Dreossi2018SemanticAD}, such as adding
stickers to a stop sign that result in misclassification \cite{Eykholt2017RobustPA}.
However, these attacks still focus on perception models disembodied from the target application, such as autonomous driving, and
%However, existing investigations on adversarial examples still focus on classification errors associated with static images and are conducted in limited experimental environments.
%~\cite{Eykholt2017RobustPA, Lu2017NONT, Dreossi2018SemanticAD}.
few efforts study such attacks deployed directly on dynamical systems, such as autonomous driving \cite{Tuncali2018SimulationbasedAT,KALRA2016182}.

\section{Modeling Framework}
\label{sec:modeling_framework}

% \begin{figure*}
%   \centering  \includegraphics[width=\textwidth]{media/fig2-flowchart.jpg}
%   \caption{Architecture overview of our simulation infrastructure including the interfaces between the CARLA simulator and the pattern generator scripts. Visualization of the camera and the third person views from one attack episode are also shown.}
%   \label{fig:overall_architecture}
% \end{figure*}

In this paper, we focus on exploring the influence of a physical adversary that successfully subverts RGB camera-based e2e driving models. We define physical adversarial examples as attacks that are physically realizable in the real world. For example, deliberately painted shapes on the road or on stop signs would be classified as physically realizable. Fig. \ref{fig:conceptual_overview}(b) displays the conceptual view of such an attack involving painting black lines. We define our adversarial examples as \textit{patterns}. To create an adversarial example that forces the e2e model to crash the vehicle, we need to choose the parameters of \textit{pattern}'s shape that maximize the objective functions that we present. This may cause the vehicle to veer into the wrong lane or go offroad, which we characterize as a successful attack. Conventional gradient-based attack techniques 
are not directly applicable, since we need to run simulations (using the CARLA autonomous driving simulator) both to implement an attack pattern, and to evaluate the end-to-end autonomous driving agent's performance.
%cannot be applied in this domain since the generated attacks requires pixel-level modifications spanning the entire input space, which is not physically realizable.

%Our first approach to the problem is to systematically explore a  relatively coarsely discretized search space. We choose regions of interest on the road where we will create attacks, and begin by \textit{drawing} a simple pattern like a thin black line with fixed width and length on those regions. Then we sweep through different positions and orientations of the pattern to see if exhaustively going through the search space finds adversarial examples that cause the vehicle to crash. 
%Finally, we apply an optimization technique to search this space more efficiently. Note that for the entirety of this research, we use the right-driving traffic system.
%Since this approach is becomes relatively ineffective as we increase the dimensionality of the search space, we present another technique for to identify effective attacks based on Bayesian optimization.

At the high level, our goal is to paint a pattern (such as a black line) somewhere on the road to cause a crash. We formalize such attacks in terms of optimizing an objective function that measures the success of the attack pattern at causing driving infractions. 
Since driving infractions themselves are difficult to optimize because of discontinuity in the objective (infraction either occurs, or not), one of our goals it to identify a high-quality proxy objective.
Moreover, since the problem is dynamic, we must consider the impact of the object we paint on the road over a sequence of frames that capture the road, along with this pattern, as the vehicle moves towards and, eventually, over the modified road segment.
Crucially, we modify the road itself, which is subsequently captured in vision, digitized, and used as input into the e2e model's controller.

To formalize, we now introduce some notation.
Let $\delta$ refer to the pattern painted on the road, and 
let $l$ denote the position on the road where we place the pattern.
%, which we, in turn, denote by $\delta$.
We use $L$ to denote the set of feasible locations at which we can position the adversarial pattern $\delta$, and $S$ the set of possible patterns (along with associated modifications; in our case, we consider either a single black line, or a pair of black lines, with modifications involving, for example, the distance between the lines, and their rotation angles).
% Let $x_l$ be the state of the track at position $l$, and $x_l +\delta$ then becomes the state of the track at this same position when the pattern $\delta$ is added to it.
% The state of the track at position $l$ is captured by the vehicle's vision system when it comes into view; we denote the frame at which this location initially comes into view by $f_l$, and let $\Delta$ be the number of frames over which the track in position $l$ is visible to the vehicle's vision system.
% Given the track state $x_l$ at position $l$, the digital view of it in frame $f$ is denoted by $y_f(x_l)$.
% Finally, we let $f_{sa}(y_f,h_f)$ denote the predicted steering angle given observed digital image corresponding to frame $f$, and prior history of frames, $h_f$.
Let $a_l$ be the state of the road at position $l$, and $a_l +\delta$ then becomes the state of the road at this same position when the pattern $\delta$ is added to it.
The state of the road at position $l$ is captured by the vehicle's vision system when it comes into view; we denote the frame at which this location initially comes into view by $F_l$, and let $\Delta$ be the number of frames over which the road in position $l$ is visible to the vehicle's vision system.
Given the road state $a_l$ at position $l$, the digital view of it in frame $F$ is denoted by $I_F(a_l)$ or simply $I_F$.
Finally, we let $\theta_F = g_{sa}(I_F)$ denote the predicted steering angle given observed digital image corresponding to frame $F$.  With this formalism established, we introduce several candidates for a proxy objective function that would quantify the success of an attack. 

\subsection{Candidate Objective Functions}
\label{subsec:candidate_objective_functions}

\subsubsection{Steering Angle Summations}
\label{subsubsec:steering_angle_summations}
First, we denote the vector of predicted steering angles during an episode \textit{with an attack} $\delta$ starting from frame $F_{l}$ to frame $F_{l + \Delta}$ as:
\begin{equation}
\label{E:steering_angles_with_attack}
    \vec{\Theta}_{\delta} = [\theta_{F_{l}}, \theta_{F_{l + 1}}, \cdots, \theta_{F_{l + \Delta}}]
\end{equation}
We define two objective functions as:
\begin{subequations}
\label{E:attack1}
\begin{align}
\mathrm{Collide\ Right:\ }&\max_{l,\delta} \sum_{i = 0}^{\Delta} \vec{\Theta}_{{\delta}_i} \label{E:attack1a}\\
\mathrm{Collide\ Left:\ }&\min_{l,\delta}  \sum_{i = 0}^{\Delta} \vec{\Theta}_{{\delta}_i} \label{E:attack1b}\\
&\mathrm{subject\ to:}\quad l \in L, \quad \delta \in S.
\end{align}
\end{subequations}
Equation \ref{E:attack1a} says that to optimize an attack that causes the vehicle to veer off towards the right and collide, we need to maximize the sum of steering angles for that particular experiment for the frames in which the pattern is in view. And similarly in Equation \ref{E:attack1b}, we need to minimize the steering sum, to make the vehicle veer left. We convert Equation \ref{E:attack1b} to a maximization problem for consistency in our search procedures that we will describe. Using Equation \ref{E:attack1} as the objective function allows us to have control over which direction we would like the car to crash. The following two metrics, the absolute steering angle difference and path deviation, lose this ability to distinguish direction-based attacks, since they are essentially L-1 and L-2 norms.

\subsubsection{Absolute Steering Angle Differences}
\label{subsubsec:absolute_steer_diff}
Again, let's denote the predicted steering angles \textit{with an attack} $\delta$ over the frames $F_l$ to $F_{l + \Delta}$ as $\vec{\Theta}_{\delta}$ as shown in Equation \ref{E:steering_angles_with_attack}. Now, let's denote the predicted steering angles \textit{without any attack} over the same frames as $\vec{\Theta}_{\text{baseline}}$. This represents an episode where no attack is added to the road (we refer to this as the baseline run) and the car travels the intended path with minimal infractions. We can now define our second candidate metric as:
\begin{subequations}
\label{E:attack2}
\begin{align}
\max_{l,\delta} ||\vec{\Theta}_{\delta} - \vec{\Theta}_{\text{baseline}}{||}_1 \label{E:attack2a}\\
\mathrm{subject\ to:}\quad l \in L, \quad \delta \in S. \label{E:attack2b}
\end{align}
\end{subequations}
Equation \ref{E:attack2a} optimizes an attack over the frames $\Delta$ that cause the largest absolute deviation in predicted steering angles with respect to the predicted steering angles when no pattern has been added to the road.

\subsubsection{Path Deviation}
\label{subsubsec:path_deviation}
First denote the $(x,y)$ position of the agent from frames $F_l$ to $F_{l + \Delta}$ \textit{with an attack} $\delta$ as:
\begin{equation}
    \vec{\text{p}}_{\delta} = [(x_l, y_l), (x_{l+1}, y_{l+1}), \cdots, (x_{l+\Delta}, y_{l+\Delta})]
\end{equation}
Define $\vec{\text{p}}_{\text{baseline}}$ as the position of the agent \textit{with no attack} added to the road over the same frames (the baseline run). 
We can optimize the path deviation from the baseline path:
\begin{subequations}
\label{E:attack3}
\begin{align}
\max_{l,\delta} ||  \vec{p}_{\delta} -\vec{p}_{\text{baseline}} {||}_2 \label{E:attack3a}\\
\mathrm{subject\ to:}\quad l \in L, \quad \delta \in S. \label{E:attack3b}
\end{align}
\end{subequations}
Similar to Equation \ref{E:attack2a}, we can use this metric to optimize deviation from the baseline route, except we are now attacking the position of the vehicle which is directly influenced by the outputs of the e2e models.

\section{Approaches for Generating Adversaries}
\label{sec:approaches_for_generating_adversaries}

We now describe our approaches for computing adversarial patterns or, equivalently, optimizing the objective functions defined above.

\subsection{Random and Grid Search}
\label{subsec:random_grid_search}

Each \textit{pattern} we generate (labeled earlier as $\delta$) can be described by a set of parameters such as length, width, and rotation angle with respect to the road. Two naive methods of finding successful attacks would be to generate a pattern through either a random or grid search (using a coarse grid) and evaluate this pattern using one of the above objective functions. Algorithm \ref{algorithm:algo1} shows this setup. The function \textit{RunScenario()} runs the simulation and returns data such as vehicle speed, predicted acceleration, GPS position, and steering angle. We use these results to calculate one of the objective functions (\textit{CalculateObjectiveFunction()}). As our goal is to maximize this metric, we use \textit{MetricsList} to store the results of the objective function at each iteration. Finally, we return the parameters that maximized our objective function.

\begin{algorithm}
\caption{Adversary Search Algorithm}
\begin{algorithmic} 
\label{algorithm:algo1}
\REQUIRE $\text{Strategy} \in \text{Random, Grid}$
\STATE $i \leftarrow 0$
\STATE $\text{MetricsList} \leftarrow [\text{ } ]$
\LOOP
\STATE $\delta_i \leftarrow  \text{GenerateAttack}(\text{Strategy})$
\STATE results $ \leftarrow $ RunScenario($\delta_i$)
\STATE $y_i \leftarrow $ CalculateObjectiveFunction(results)
\STATE $\text{MetricsList}$.append($y_i$)
\STATE $i \leftarrow i + 1$
\ENDLOOP
\RETURN $\arg \max \text{MetricsList}$
\end{algorithmic}
\end{algorithm}

\subsection{Bayesian Optimization Search Policy}
\label{subsec:bayesian_optimization}

Algorithm \ref{algorithm:algo1} works well when the number of parameters for $\delta$ are relatively small. 
%For a larger pattern search space, we turn to an optimization strategy that infers from previously generated patterns and their objective function values. 
For a larger pattern space, and to enable us to explore the space more finely, we turn to Bayesian Optimization, which is designed for optimizing an objective function that is expensive to query without requiring gradient information \cite{BO}. It has been shown that Bayesian Optimization (BayesOpt) can be useful for optimizing expensive functions in various domains such as hyper-parameter tuning, reinforcement learning, and sensor calibration \cite{1807.02811,garnett,1805.04748,1502.05700}. In our case, since we use an autonomous driving simulator, it is \textit{expensive} to run a simulation with a generated attack in order to find, for example, the sum of steering angles as shown in Equation \eqref{E:attack1}. 
On average, one episode takes between 20 to 40 seconds depending upon the scenario; consequently, it is important for the optimization to be sample efficient.
%In order to effectively explore the search space of the physical adversaries we intend to generate, we employ Bayesian Optimization as one of our search policies. 
%Bayesian Optimization  is able to  optimize an objective function that is expensive to query \cite{BO}. 

-At the high level, our goal is to generate physical adversaries that successfully attack e2e autonomous driving models, where a successful attack can be quantified as trying to maximize some objective function $f(\delta)$. 
Our goal, therefore, is to find a physical attack, $\delta^*$, such that:
\begin{equation}
    \delta^* = \arg \max_\delta f(\delta),
\end{equation}
where $\delta^* \in \mathcal{R}^d$ and $d$ is the number of parameters of the physical attack. 
%We expedite our search procedure by using Bayesian Optimization. 
We first assume that the objective $f$ can be represented by a Gaussian Process, which we denote by $GP(f,\mu(\delta), k(\delta, \delta'))$ with a mean function of $\mu(\delta)$ and a covariance function $k(\delta, \delta')$ \cite{Rasmussen06gaussianprocesses}. We assume the prior mean function to be $\mu(\delta) = \mathbf{0}$ and the covariance function to be the Mat\'ern $5/2$ kernel:
\begin{equation}
    k(\delta, \delta') = \left(1+\frac{\sqrt{5} r}{\ell}+\frac{5 r^{2}}{3 \ell^{2}}\right) \exp \left(-\frac{\sqrt{5} r}{\ell}\right),
\end{equation}
where $r$ is the Euclidean distance between the two input points, $||\delta - \delta'||_2 $, and $\ell$ is a scaling factor optimized during simulation run-time. Let's suppose that we have already generated several adversaries and evaluated our objective function $f$ for each of these adversaries. We can denote this dataset as $\mathcal{D} =  \{(\delta_1, y_1), \cdots, (\delta_{n-1}, y_{n-1}) \}$. Therefore, if we would like to sample our function $f$ at some point along the input space $\delta$, we would obtain some posterior mean value $\mu_{f|\mathcal{D}}(\delta)$ along with a posterior confidence or standard deviation value of $\sigma_{f|\mathcal{D}}(\delta)$.  As noted earlier, our objective function $f$ is expensive to query. When we use Bayesian optimization to find the parameters that define our next adversary $\delta_{n}$, we instead maximize a proxy function known as the acquisition function, $u(\delta)$. Compared to the objective function, it is trivial to maximize the acquisition function using an optimizer such as the L-BFGS-B algorithm with a number of restarts to avoid local minima. In our case, we utilize the Expected Improvement (EI) acquisition function. Given our dataset, $\mathcal{D}$, we first let $y_{\text{max}}$ be the highest objective function value we have seen so far. The EI can be evaluated at some point $\delta$ as:
\begin{equation}
    u(\delta) = \mathrm{E}[\max (0, f(\delta) - y_{\text{max}})].
\end{equation}
Given the properties of a Gaussian Process, this can be written in closed form as follows:
\begin{gather}
    z = \frac{\mu_{f|\mathcal{D}}(\delta) - y_{\text{max}}}{\sigma_{f|\mathcal{D}}(\delta)}; \\
    u(\delta) = (\mu_{f|\mathcal{D}}(\delta) - y_{\text{max}}) \Phi\left(z \right) + \sigma_{f|\mathcal{D}}(\delta) \phi\left(z \right),
\end{gather}
where $\Phi$ and $\phi$ are the cumulative and probability distribution functions of the Gaussian distribution. Effectively, the first term in the above acquisition function leads to exploiting information from previously generated adversaries to generate parameters for $\delta_n$ while the second term prefers exploring the input space of the adversary parameters. Given this setup, Algorithm~\ref{algorithm:algo2} presents a Bayesian Optimization approach for generating and searching for adversarial patterns.
\begin{algorithm}

\caption{Bayesian Adversary Search Algorithm}
\begin{algorithmic} 
\label{algorithm:algo2}
\STATE $i \leftarrow 0$
\STATE $\text{MetricsList} \leftarrow [\text{ } ]$
\LOOP
\STATE $\delta_i \leftarrow  \arg \max u(\delta)$
\STATE results $ \leftarrow $ RunScenario($\delta_i$)
\STATE $y_i \leftarrow $ CalculateObjectiveFunction(results)
\STATE $\text{MetricsList}$.append($y_i$)
\STATE Update Gaussian Process and $\mathcal{D}$ with $(\delta_i, y_i)$
\STATE $i \leftarrow i + 1$
\ENDLOOP
\RETURN $\arg \max \text{MetricsList}$
\end{algorithmic}
\end{algorithm}
In this algorithm, the Gaussian process is updated in each iteration, and the acquisition function reflects those changes. An initial warm-up phase where the adversary parameters are chosen at random and the simulation is queried for the objective function is used for hyper parameter tuning.

\section{Experimental Methodology}
\label{sec:experimental_methodology}

\begin{figure*}
  \centering  \includegraphics[width=\textwidth]{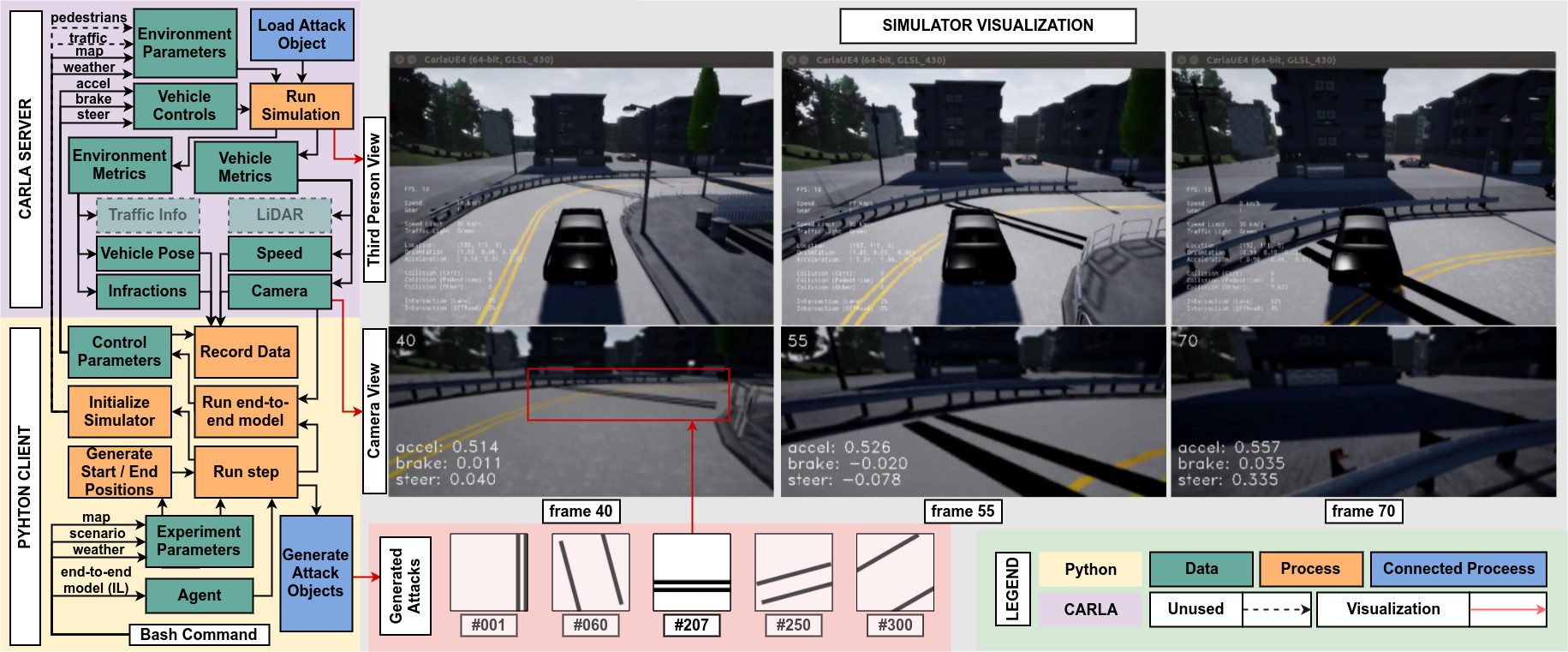}
  \caption{Architecture overview of our simulation infrastructure including the interfaces between the CARLA simulator and the pattern generator scripts. Visualization of the camera and the third person views from one attack episode are also shown.}
  \label{fig:overall_architecture}
\end{figure*}

This section introduces the various building blocks that we use to perform our experiments. Fig.~\ref{fig:overall_architecture} shows the overall architecture of our experimentation method, including the CARLA simulator block, the python client block, and how they communicate with each other to generate and test the attack patterns on the simulator.

\subsection{Autonomous Vehicle Simulator}
\label{subsec:autonomous_vehicle_simulator}

Autonomous driving simulators are often used to test autonomous vehicles for the sake of efficiency and safety \cite{Shah2017AirSimHV, Fan2018BaiduAE, Tian2018DeepTestAT, requirement_driven_test_generation}. After testing popular autonomous simulators \cite{deepdrive, baidu_apollo, nvidia_driveworks, airsim}, we chose to run our experiments on the CARLA \cite{Dosovitskiy2017CARLAAO} (CAR Learning to Act) autonomous vehicle simulator, due to its feature-set and ease of source code modification . With Unreal Engine 4 \cite{UnrealEngine4} as its backend, CARLA has sufficient flexibility to create realistic simulated environments, with a robust physics engine, lifelike lighting, 3D objects including roads, buildings, traffic signs, vehicles and pedestrians. Fig.~\ref{fig:overall_architecture} shows how the simulator looks in the third person view. It allows us to acquire sensor data like the camera image for each frame (camera view), vehicle measurements (speed, throttle, steering angle and brake) and other environmental metrics like how the vehicle interacts with the environment in the form of infractions and collisions. Since we use e2e models that use only the RGB camera, we disable the LiDAR, semantic segmentation, and depth cameras. Steering angle, throttle and brake parameters are the primary control parameters to drive the vehicle in the simulation. CARLA (v0.8.2) comes with two maps: a large training map and a smaller testing map which are used for training and testing the e2e models respectively. CARLA also allows the user to run experiments under various weather conditions like sunset, overcast and rain, which are determined by the client input. To keep consistent frame rate and execution time, we run CARLA using a fixed time-step.

\subsection{End-to-end Driving Models}
\label{subsec:end_to_end_driving}

The CARLA simulator comes with two trained end-to-end models: Conditional Imitation Learning (IL) \cite{Codevilla2018EndtoEndDV} and Reinforcement Learning (RL) \cite{Dosovitskiy2017CARLAAO}. Their commonality ends at using the camera image as the input to produce output controls that include steering angle, acceleration, and brake. The IL model uses a trained network which consists of demonstrations of human driving on the simulator. In other words, the IL model tries to mimic the actions of the expert from whom it was trained  \cite{1801.06503}.  The IL model's structure comprises of a conditional, branched neural architecture model where the \textit{conditional} part is a high-level command given by the CARLA simulator at each frame. This high-level command can be \textit{left, right or straight at an intersection, and lane follow when not at an intersection}.

% \begin{figure}
%   \centering  \includegraphics[width=\columnwidth]{media/model_IL.png}
%   \caption{The Imitation Learning (IL) model consists of a conditional, branched neural network trained to mimic the actions of a human driver within the simulation. The branching is conditioned on the current CARLA high-level command (as specified by \textbf{c} in the figure).\cite{Codevilla2018EndtoEndDV} }
%   \label{fig:modelIL}
% \end{figure}

At each frame, the image, current speed, and high-level command are used as inputs to the branched IL network to directly output the controls of the vehicle. Therefore, each branch is allocated a sub-task within the driving problem (making a decision to cross an intersection or following the current lane). The RL model uses a trained deep network based on a rewards system, provided by the environment based on the corresponding actions, without the aid of human drivers. More specifically, for RL, the asynchronous advantage actor-critic (A3C) algorithm was used. It is worth mentioning that the IL model performed better than the RL model in untrained (test) scenarios \cite{Dosovitskiy2017CARLAAO}.  Because of this, we focus our research primarily on attacking the IL model. 

\subsection{Physical Adversary Generation}
\label{subsec:physical_adversary_generation}

\subsubsection{Unreal Engine}
\label{subsubsec:unreal_engine}

To generate physically realizable adversaries in a systematic manner,
we modify CARLA's source code. The CARLA simulator (v0.8.2) does not allow spawning of objects into the scene which do not already exist in the CARLA blueprint library (which includes models of vehicles, pedestrians, and props). With the Unreal Engine 4 (UE4), we create a new \textit{Adversarial Plane Blueprint}, which is a $200 \times 200$ pixel plane or canvas with a dynamic UE4 material, which we can overlay on desired portions of the road. The key attribute of this blueprint is that it reads a generated attack image (a .png file) and places it within CARLA in real time. Hence this blueprint has the ability to continuously read an image via an HTTP server. The canvas allows the use of images with an alpha channel which allows attacks which are partly transparent, like the one shown in Fig. \ref{fig:jsa_attack_patterns}.  
Then, we clone the two maps that are provided by CARLA and choose regions of interest within each of them where attacks spawn. Some interesting regions are at the turns and intersections. We place the \textit{Adversarial Plane Blueprint} canvas in each of these locations. When CARLA runs, an image found on the HTTP server gets overlaid on each canvas. Finally, we compile and package this modified version of CARLA. Hence we are able to place physical attacks within the CARLA simulator.

\subsubsection{Pattern Generator Library}
\label{subsubsec:pattern_generator_library}

\begin{table}[]
\centering
\caption{Different types of attacks and their respective parameters and constraints. var - variable, const - constant, NA - Not Applicable}
\begin{tabular}{ccccc}
\hline
\multicolumn{1}{l}{} & \multicolumn{4}{c}{\textbf{Attack Type}}                                            \\
\textbf{params}      & \textbf{Single Line} & \textbf{Double Line} & \textbf{Two Lines} & \textbf{N-Lines} \\ \hline
\textbf{\# lines}    & 1                    & 2                    & 2                  & N                \\
\textbf{position}    & var                  & var                  & var                & var              \\
\textbf{rotation}    & var                  & var                  & var                & var              \\
\textbf{length}      & const                & const                & const              & var              \\
\textbf{width}       & const                & const                & const              & var              \\
\textbf{gap}         & NA                   & var                  & NA                 & NA               \\
\textbf{color}       & const                & const                & const              & var              \\
\textbf{opacity}     & const                & const                & const              & var              \\ \hline
\textbf{dimensions}  & 2                    & 3                    & 4                  & N * 6            \\ \hline
\end{tabular}
\newline
\label{tab:attack_parameters}
\end{table}

\begin{figure}
  \centering  \includegraphics[width=\columnwidth]{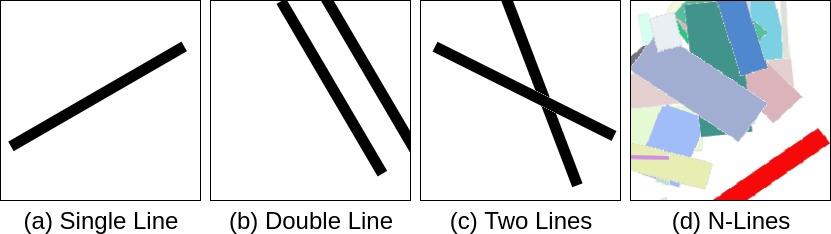}
  \caption{ Attack Generator Capabilities. (a) shows the most basic attack which is a single line. (b) and (c) show attacks using two lines, but (b) has a constraint that the lines always need to be parallel, (d) shows the ability of the generator to generate N number of lines with various shapes and color. }
  \label{fig:jsa_attack_patterns}
\end{figure}

We built a pattern generator that creates different kinds of shapes as shown in Fig. \ref{fig:jsa_attack_patterns} using the pattern parameters (Table \ref{tab:attack_parameters}). For the pattern generator, we explore parameters like the position, width, and rotation of the line(s). We sweep the position and rotation from 0-200 pixels and 0-180 degrees respectively to generate variations of attacks. Similarly, we create a more advanced pattern which involves two parallel black lines called the \textit{double-line} pattern as described in Table \ref{tab:attack_parameters}. It comprises of the previous parameters, namely, position, rotation, and width, with the addition of a new gap parameter which is the distance between the two parallel lines.  Lastly, we remove the parallel constraint on double lines to increase the search space of the attacks while preserving simplicity.  Fig.~\ref{fig:overall_architecture} shows some examples of the generated double line patterns which can be seen overlaid on the road in frames 55 and 70. 

Additionally, our library has the ability to read a dictionary object containing the number of lines, the parameters (position, rotation, width, length and color) for each line, and produce a corresponding attack pattern as shown in Fig. \ref{fig:jsa_attack_patterns} (d). Once the pattern is generated, it is read via the HTTP server and is placed within the Carla simulator.	

\subsubsection{OpenAI-Gym Environment for Carla}
\label{subsubsec:open_ai_gym_carla}

Since CARLA runs nearly in real-time, experiments take a long time to run. In order to efficiently run simulations with our desired parameters, we convert the CARLA setup to an OpenAI-Gym environment \cite{1606.01540}. While the OpenAI-Gym framework is primarily used for training reinforcement learning models, we find the format helpful as we are able to easily run the simulator with a set of initial parameters like the task (straight, right, left), the map, the scene, the end-to-end model and the desired output metric (eg. average infraction percent for that episode). With this set up, we are able to use an optimizer to generate an attack with a set of defined constraints, run an episode and get the resulting output metric. 

\subsection{Experiment Setup and Parallelism}
\label{subsec:experiment_setup}

To ensure a broad scope to test the effectiveness of the different attacks in various settings, we conduct experiments by changing various environment parameters like the maps (training map and testing map), scenes, weather (clear sky, rain, and sunset), driving scenarios (straight road, right turn, and left turn), e2e models (IL and RL) and the entire search space for the patterns.  Here, we describe the six available driving scenarios for CARLA:
\begin{enumerate}
    \item \textit{Right Turn}: the agent follows a lane that smoothly turns 90 degrees towards the right.
    \item \textit{Left Turn}: the agent follows a lane that smoothly turns 90 degrees towards the left.
    \item \textit{Straight Road}: the agent follows a straight path.
    \item \textit{Right Intersection}: the agent takes a right turn at an intersection.
    \item \textit{Left Intersection}: the agent takes a left turn at an intersection.
    \item \textit{Straight Intersection}: the agent navigates straight through intersection.
\end{enumerate}

% Our experiment follows the setup shown in Fig. \ref{fig:overall_architecture}. Here, the function GenerateAttack() uses an optimizer (e.g. grid search) to produce attack patterns that are then placed within the CARLA environment. The recorded value of the metric is used to quantify the strength of an attack and to optimize the search method.

% \begin{algorithm}
% \caption{Adversary Search Algorithm}
% \begin{algorithmic} 
% \label{algorithm:adversary_search_algorithm}
% \REQUIRE $\text{Model} \in \text{IL, RL}$
% \REQUIRE $\text{Task} \in \text{Straight Road, Left Corner, Right Corner}$
% \STATE $\text{Attack} \leftarrow \text{NULL} $
% \STATE $\text{Baseline Steers}\leftarrow \text{RunScenario(Model, Task)}$
% \LOOP 
% \STATE $\text{Attack} \leftarrow \text{ GenerateAttack()}$
% \STATE $\text{Attack Steers} \leftarrow \text{RunScenario(Model, Task)}$
% \STATE $\text{Metric} \leftarrow \sum^{\Delta}_{\tau = 0} \text{Attack Steers}$
% \STATE $\text{Record Metric}$
% \ENDLOOP
% \end{algorithmic}
% \end{algorithm}
% 

We choose the baseline scenarios (no attack) where the e2e models drive the vehicle with minimal infractions. We run the experiments at 10 frames per second (fps) and collect the following data for each camera frame (a typical experiment takes between 60 to 100 frames to run): camera image from the mounted RGB camera, vehicle speed, predicted acceleration, steering and brake, percentage of vehicle in the wrong lane, percentage of vehicle on the sidewalk (offroad), GPS position of the vehicle, and collision intensity. Fig. \ref{fig:overall_architecture} shows this dataflow which is sufficient to assess the ramifications of a particular attack in an experiment. 

To search the design space thoroughly, we build a CARLA docker which allows us to run as many as 16 CARLA instances simultaneously, spread over 8 RTX GPUs~\cite{RTX}. 
% We chose two different scenes for the three tasks in each of the training and testing maps. Since we would be running 374 combinations of double lines, and 111 combinations of single line patterns over the 72 different scenarios,
%Next we discuss how we found successful attacks against the different driving scenarios and end-to-end models.

\section{Experimental Results}
\label{sec:experiment_results}
Through experimentation, we demonstrate the existence of conspicuous physical adversaries that successfully break the e2e driving models. These adversaries do not need to be subtle or sophisticated modifications to the scene. Although they can be distinguished and ignored by humans drivers with ease, they  effectively cause serious traffic infractions against the e2e  driving models we evaluate.

\subsection{Simple Physical Adversarial Examples}
\label{subsec:discovery_physical_adversaries}

\subsubsection{Effectiveness of Attacks}
\label{subsubsec:summary_of_attacks}

\begin{figure}
  \centering  \includegraphics[width=\columnwidth]{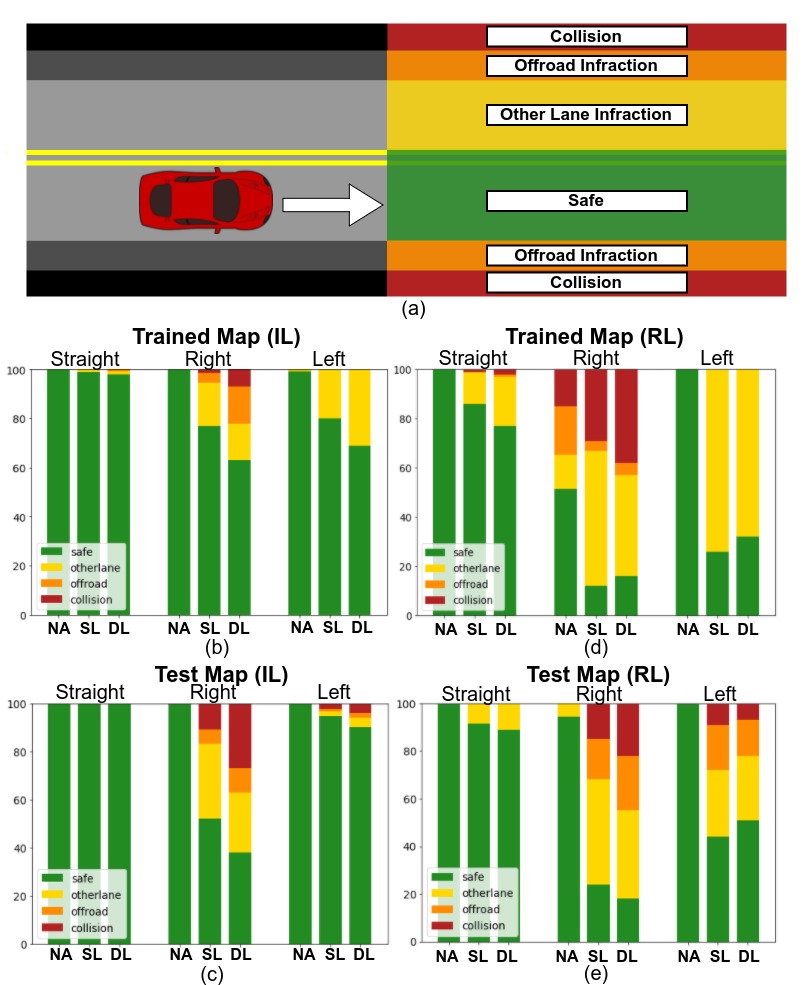}
  \caption{Comparison of the infractions caused by different patterns. (a) Driving Infraction regions; (b)(c) Infraction percentages for IL; (d)(e) Infraction percentages for RL; NA - No Attack, SL - Single Line pattern, DL - Double Lines pattern; Straight - Straight Road Driving, Right - Right Turn Driving, Left - Left Turn Driving}
  \label{fig:result01}
\end{figure}

To begin, we generated two types of adversarial patterns: single line (with varying positions and rotation angles), and double lines (with varying positions, rotation angles, and distance between the lines). In Fig.~\ref{fig:result01}(a), we define different safety regions of the road in ascending order of risk. We start with the vehicle's own lane (safe region), the opposite lane (unsafe), offroad/sidewalk (dangerous) and regions of collisions (very dangerous) past the offroad region. Fig.\ref{fig:result01}(b)(c)(d)(e) shows that by sweeping through the three scenarios (straight road driving, right turn driving, left turn driving) with the single and double line patterns, for both the training map and testing maps, we see that some patterns cause infractions. Here we use a naive grid search approach to traverse the search space with the \textit{Steering Sum optimization metric} defined in Equation \ref{E:attack1a}.
% We call these patterns that successfully cause the e2e models to make a mistake, adversarial examples. 
First, we observe the transfer-ability of adversaries since some of our generated adversarial examples cause both IL (Fig.\ref{fig:result01}(b)) and RL (Fig.\ref{fig:result01}(d)) models to produce infractions. Second, attacks are more successful against the RL model than the IL model. Additionally, we notice that the double line adversarial examples cause more severe infractions than their single line counterparts. Lastly, we observe that \textit{Straight Road Driving} and \textit{Left Turn Driving} are more resilient to attacks that cause stronger infractions. 
%In the next section, we discuss how to circumvent the limitations of the current metric and optimization techniques to attack more robust scenarios, like \textit{Straight Driving}. Later, we analyze more thoroughly the qualities of attack patterns that allow high infraction attacks against certain scenarios such as \textit{Right Corner Driving}.

\subsubsection{Analysis of Attack Objectives}
\label{subsubsec:analysis_of_attack_objectives}

\begin{figure}
  \centering  \includegraphics[width=\columnwidth]{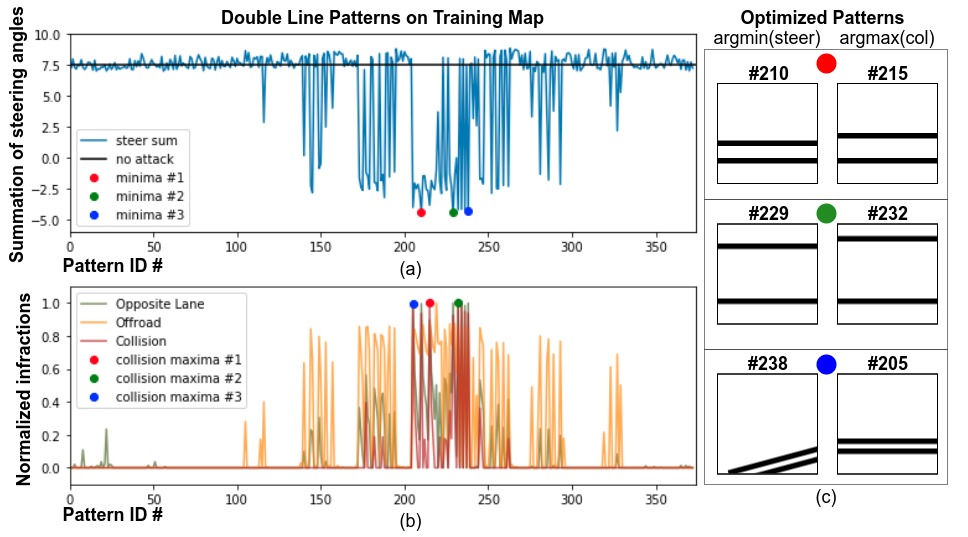}
  \caption{Adversary against "Right Turn Driving". (a) Adversarial examples significantly changes the steering control. (b) Some patterns cause minor infractions whereas others cause level 3 infractions. (c) The patterns that cause the minimum steering sum and maximum collisions look similar.}
  \label{fig:result02}
\end{figure}

To find the optimal adversary which produce infractions like collisions for the case of \textit{Right Turn Driving} scenario, the optimizer has to find a pattern that maximizes the first candidate objective function: the sum of steering angles as hypothesized in Equation \ref{E:attack1}. A positive steering angle denotes steering towards the right and a negative steering angle implies steering towards the left. Fig.~\ref{fig:result02}(a)(b) show the sum of steering angles and the sum of infractions respectively, for each of the 375 combinations of double line patterns. The infractions are normalized because collision data is recorded in SI units of intensity [$kg \times m/s$], whereas the lane infractions are in percentages of the vehicle area in the respective regions. Fig. \ref{fig:result02} also shows the three lowest points (minima) for the steering sum and the three highest points (maxima) for the collisions plot. In Fig.~\ref{fig:result02}(c), we use the \textit{argmin} and \textit{argmax} on the set of attacks to observe the shapes of the corresponding adversarial examples for both the steering sum and infraction results. We observe that the \textit{patterns that minimize the sum of steering angle and correspondingly maximize the collision intensity are very similar.}  
Thus, the objective based on maximizing or minimizing steering angles is clearly yielding valuable information for the underlying optimization problem.
However, this does not mean that it's the best objective, among the three choices we considered above.
We explore this issue in greater depth in the next subsection, as we move towards studying more complex attacks using Bayesian optimization.
%In the following subsection, we demonstrate that BayesOpt allows us to efficiently search an attack space with a higher number of input parameters while preserving the simplicity of the attack. Additionally, we show that some candidate objective functions are better suited for BayesOpt.

\subsection{Exploration of Large Design Spaces}
\label{subsec:exploration_of_large_design_spaces}

\begin{figure*}
  \centering   \includegraphics[width=1.0\textwidth]{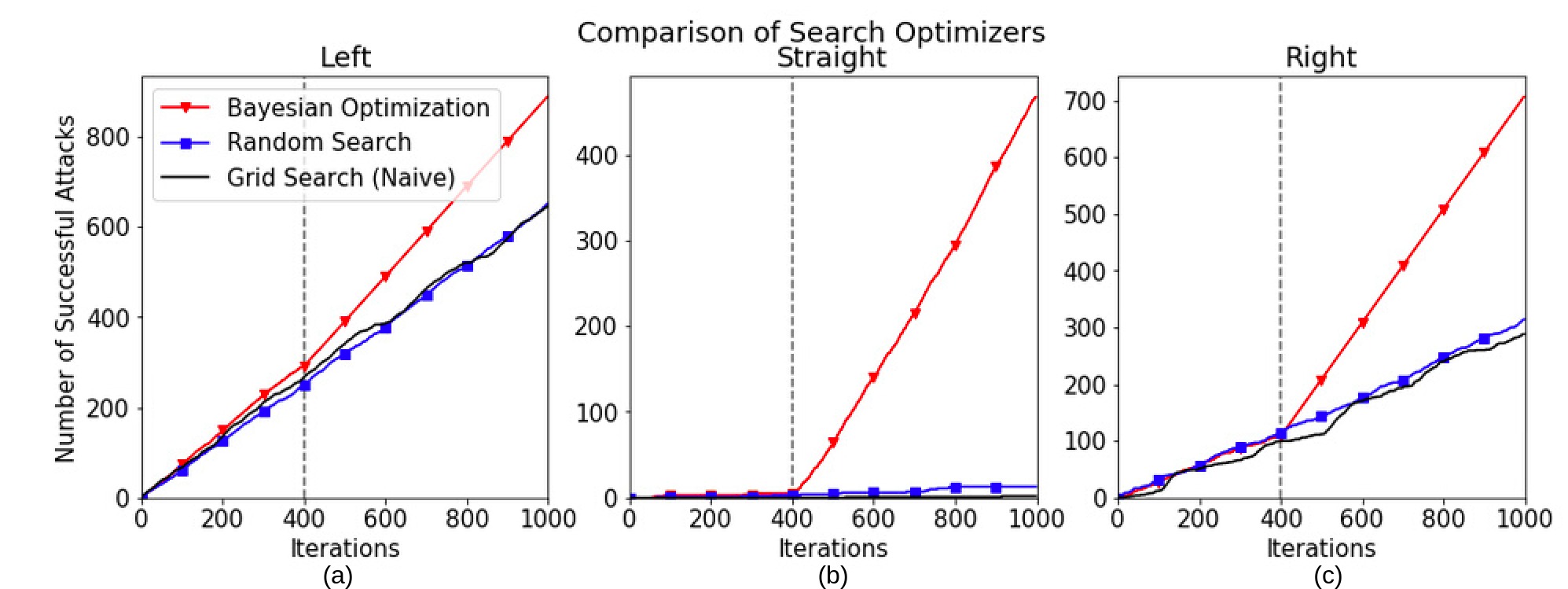}
  \caption{A comparison of different search algorithms for generating successful attacks. In each driving scenario: Left Turn (a) , Straight Road (b), and Right Turn (c) Driving, the Bayesian approach not only finds more unique, successful adversaries in the same number of iterations but also finds these attacks at a faster rate. BayesOpt randomly samples the adversary search space for the first 400 iterations (shown before the dashed line) to tune the hyper-parameters of the kernel function. After these randomly sampled points, BayesOpt utilizes an acquisition function to sample the search space. While a dense grid search would eventually find at the least the same number of attacks as BayesOpt, we constrain our experiments to 1000 iterations given our computational resources. }
  \label{fig:jsa_optimizer_comparison}
\end{figure*}

\begin{table*}[]
\centering
\caption{Comparison of Candidate Objective Functions as listed in Section III (in \%). $\sum$ st. angles - sum of steering angles, abs. st. diff. - absolute steering difference}
\label{tab:metric_comparison}
\begin{tabular}{ccccclcccclcccc}
\hline
\textbf{} & \multicolumn{4}{c}{Left} &  & \multicolumn{4}{c}{Straight} &  & \multicolumn{4}{c}{Right} \\
Metric & safe & collision & offroad & opp. lane &  & safe & collision & offroad & opp. lane &  & safe & collision & offroad & opp. lane \\ \hline
$\sum$ st. angles & 18.2 & 0 & 0 & 81.8 &  & 99 & 0 & 0 & 1 &  & 72.2 & 9.5 & 13.8 & 24.5 \\
path deviation & 64.6 & 0 & 0 & 35.4 &  & 23.8 & 2.5 & 2.8 & 76.2 &  & 57.2 & 24.0 & 28.3 & 40.2 \\
abs. st. diff. & \textbf{0.2} & 0 & 0 & \textbf{99.8} &  & \textbf{22.7} & \textbf{7.5} & \textbf{9.3} & \textbf{77.3} &  & \textbf{0} & \textbf{95.2} & \textbf{99.2} & \textbf{100} \\ \hline
\end{tabular}
\end{table*}

% \begin{figure}
%   \centering   \includegraphics[width=\columnwidth]{media/jsa_diverse_attack.jpg}
%   \caption{A suite of attacks for Right Corner Driving (a), Left Corner Driving (b), and Straight Road Driving (c). While severe infractions were found for both Right Corner and Straight Road Driving, Left Corner Driving was more resilient to double black line (the lines did not have to be parallel) attacks. It is important to note that for each case above, the \textit{Lane Follow} branch of the IL e2e model was being attacked.}
%   \label{fig:jsa_diverse_attacks}
% \end{figure}

In Fig. \ref{fig:result01}, we observe that when we switch from a \textit{Single Line} attack (with 2 dimensions) to a \textit{Double Line} attack (with 3 dimensions), in most cases, there is a significant increase in the number of successful attacks. 
It is reasonable to assume that as we increase the number of degrees of freedom in the attack, it should be possible to also increase the success rate. 
We lend further support to this intuition by considering an attack called the \textit{Two Line} attack, shown in Fig. \ref{fig:jsa_attack_patterns}(c), with 4 dimensions by removing the constraint that the two lines must be parallel.
%on the Double Line attack which has 4 dimensions as shown in 
As shown in Fig. \ref{fig:result01}, attack success rates increase considerably compared to the more restricted attack.

However, increasing the dimensionality of the attack search space makes grid search impractical.
For example, the Single Line attack with grid search requires around 375 iterations to sweep the search space at a 20 pixel resolution. Preserving the same parameter resolution (or precision) would require 1440 iterations for Double Lines, and 12,960 iterations for the Two Line attack. Naive search would require around 45 days to sweep through the search space for a \textit{single scenario} on a modern GPU. Additionally, using a sparser resolution for the attack parameters means that we would not find potential attacks which can only be found at a higher resolution. 

%To solve this problem, we look at other state of the art optimization techniques which suit problems which have expensive target functions. One of the popular methods is 
We address this issue by adopting the Bayesian Optimization framework (BayesOpt) for identifying attack patterns (introduced in Section \ref{subsec:bayesian_optimization}). This requires a change in our search procedure as shown in Algorithm 2. In short, it uses the prior history of the probed search space to suggest the next probing point. 

Fig. \ref{fig:jsa_optimizer_comparison} shows the comparison between the 3 optimization techniques we employ for the straight, left-turn, and right-turn scenarios. We see that for all three cases, BayesOpt outperforms the naive grid search and the random search methods. In Fig. \ref{fig:jsa_optimizer_comparison}, BayesOpt uses 400 initial random points to sample the search space and subsequently samples 600 optimizing points. Hence, we observe that for the first 400 iterations, BayesOpt follows closely with random search, and after probing those initial random points, we observe a significant increase in the number of successful attacks. 
%We show an effective improvement over the naive search in all cases using the Bayesian approach. 
%As hypothesized earlier, we observe significantly larger ratio of successful attacks to unsuccessful attacks as compared to the naive results which uses a lower dimensional attack space as shown in Fig. \ref{fig:result01}.

Because we observe many more successful attacks against the Left and Right Turn scenarios as compared to the Straight Scenario, Fig. \ref{fig:jsa_optimizer_comparison} further supports our notion that driving straight is harder to attack as compared to the right and left turn scenarios.

Equipped with BayesOpt, we now systematically evaluate the relative effectiveness of the different objective functions.
%to further improve the effectiveness of this optimizer. 
%Section \ref{subsec:candidate_objective_functions} introduces three candidate metrics which are potential proxies for the objective function. 
%An ideal metric is one where the infractions are directly maximized. But since the infraction score is a causal metric, we evaluate metrics that can closely approximate this function which use parameters of the vehicle that are available in real time.
Table \ref{tab:metric_comparison} shows the infractions caused by each of the objective functions (path deviation, sum of steering angles, and absolute difference in steering angles with the baseline). For Left Turn, Straight Road, and Right Turn Driving, we list the percentage out of 600 simulation runs using BayesOpt that were safe, incurred collisions, off road infractions, or opposite lane infractions. We observe that the absolute difference in steering angles with respect to the baseline run is the strongest metric when coupled with BayesOpt to discover unique, successful attacks. 
While the most natural metric would seem to be \textit{steering sum}, it is in practice considerably less effective than maximizing absolute difference in the steering angle.
%We observe that the metric we chose initially (\textit{steering sum}) does not perform as well as we expect. We go more in depth into the issue of using steering sum in the Section \ref{subsec:importance_of_selecting_objective_function}. 
The \textit{path deviation} objective function performs well in \textit{right turn} and \textit{straight scenarios}, but fails to find optimal attacks in the \textit{left turn driving} scenario. Overall it still under-performs when compared to the absolute steering difference objective function.

\subsection{Importance of Selecting a Reliable Objective Function}
\label{subsec:importance_of_selecting_objective_function}

\begin{figure}
  \centering  \includegraphics[width=1.0\columnwidth]{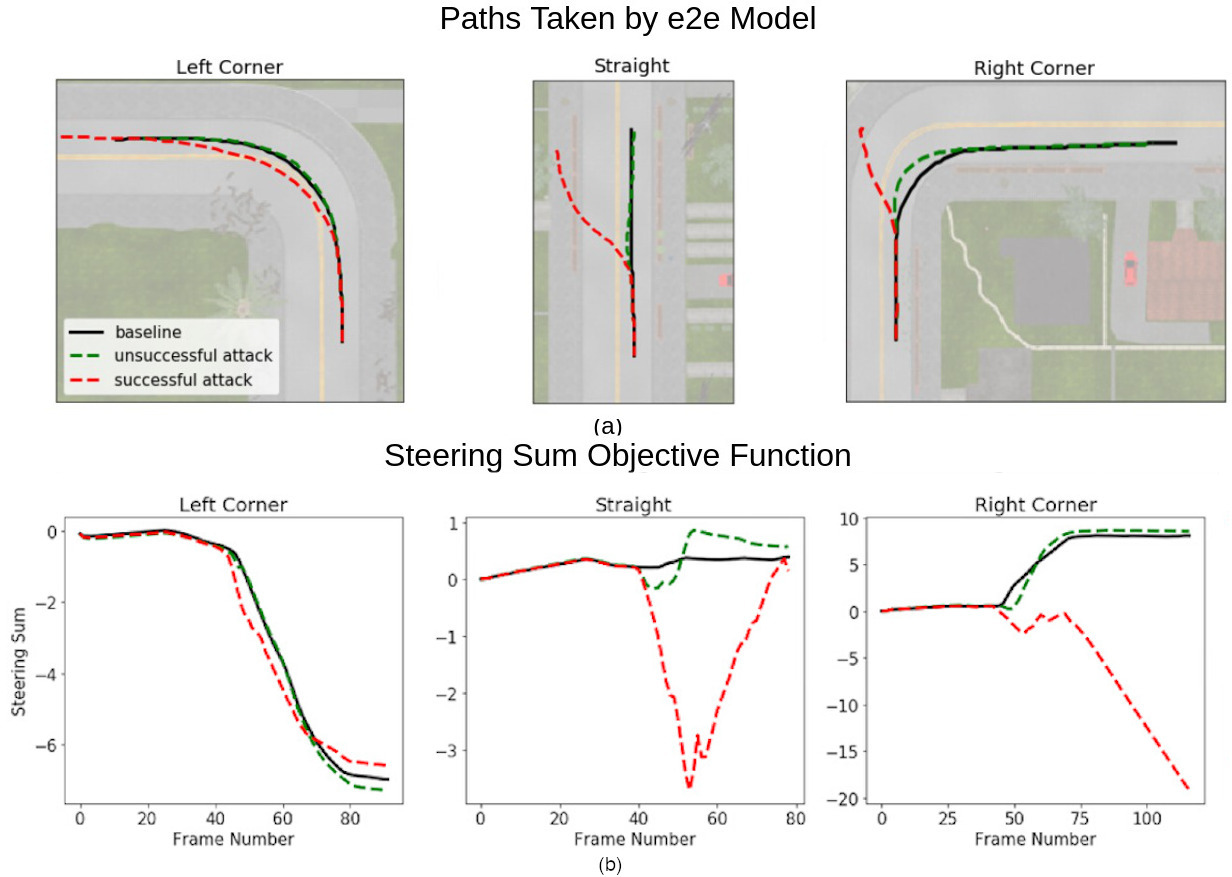}
  \caption{Paths taken by e2e model in Left Turn, Straight Road, and Right Turn Driving with no attack (baseline), an unsuccessful attack, and a successful attack (a). Cumulative sum of steering angles for each scenario (b). While the successful attack is able to cause the e2e agent to incur an infraction or collision in each scenario, the steering sum metric is unable to capture distinguish between the successful and unsuccessful attack in two of the three scenarios.}
  \label{fig:jsa_discussion_objective_function}
\end{figure}

In Section \ref{subsec:exploration_of_large_design_spaces}, we evaluated three different objective functions: \textit{path deviation}, \textit{sum of steering angles} and \textit{absolute steering difference}. We observed that the choice of the right objective function is crucial for success, and this choice is not necessarily obvious.

%The \textit{Sum of Steering Angles} and \textit{Path Deviation} objective functions were chosen out of mathematical intuition. \textit{Sum of Steering Angles} as detailed in Section \ref{subsubsec:steering_angle_summations}, uses the simple logic that the larger the sum of steering angles, the more the vehicle moves away from the road and to the right, hence stronger the attack. 
Most surprisingly, perhaps, we found that the objective that uses the steering angles to guide adversarial example construction is not the best choice, even though it is perhaps the first that comes to mind, and one used in prior work~\cite{Tian2018DeepTestAT}.
We now investigate why this choice of the objective can fail. 

Fig. \ref{fig:jsa_discussion_objective_function} shows three driving scenarios (left turn, driving straight, and right turn) respectively. Fig. \ref{fig:jsa_discussion_objective_function}(a) shows the paths taken by the vehicle for 3 cases: a \textit{baseline} case where there is no attack, an \textit{unsuccessful attack} case where an attack pattern does not cause the e2e model to deviate significantly from the \textit{baseline} path, and a \textit{successful attack} case where an attack causes a large deviation resulting in an infraction. Fig. \ref{fig:jsa_discussion_objective_function}(b) shows the sum of steering angles for each of the corresponding cases in Fig. \ref{fig:jsa_discussion_objective_function}(a). Note that for \textit{Left Turn Driving}, we try to maximize Eq.~\eqref{E:attack1a}, which is to collide to the right, and for \textit{Straight Driving} and \textit{Right Turn Driving}, we maximize Eq.~\eqref{E:attack1b}, which is to collide to the left. For the \textit{right turn driving} scenario, we observe that there is indeed a large difference between the steering sums for a strong attack and a weak attack, but in the other two scenarios, we notice that the baseline, unsuccessful attack and successful attack have very similar steering sums. Hence, the optimizer has a difficult time distinguishing between an unsuccessful and successful attack. In \textit{straight driving scenario}, we see that the steering sum for a successful attack begins increasing and then sharply decreases, even though the vehicle has deviated significantly from the baseline path. This is due to the  ability of the IL e2e model to recover in this case, resulting from data augmentation at training time where initial position of the car was randomly perturbed. The \textit{sum of steering angles} objective function is unable to capture this behavior.
For the case of \textit{left turn driving}, we discover that the successful attack not only causes a change in steering angle, but also a change in throttle, resulting in the vehicle speeding up and reaching a position further along the baseline path, which opens up new possibilities for generating attacks as well as causing new  types of infractions.

The \textit{absolute steering difference} mitigates the above  issues by summing up the absolute steering differences between the baseline and attack cases. This allows the objective function to counteract the recovery ability of the e2e models.  However, we do lose the ability to directly control the direction towards which we desire the vehicle to crash. 
%Of course, there may exist alternative objective functions that we have not considered which perform significantly better than the \textit{absolute steering difference}, (for example, taking the weighted sum of two different objective functions) which is open to further research.

\subsection{Vehicle Hijacking Attacks}
\label{subsec:case_study_hijacking}

\begin{figure}
  \centering  \includegraphics[width=0.9\columnwidth]{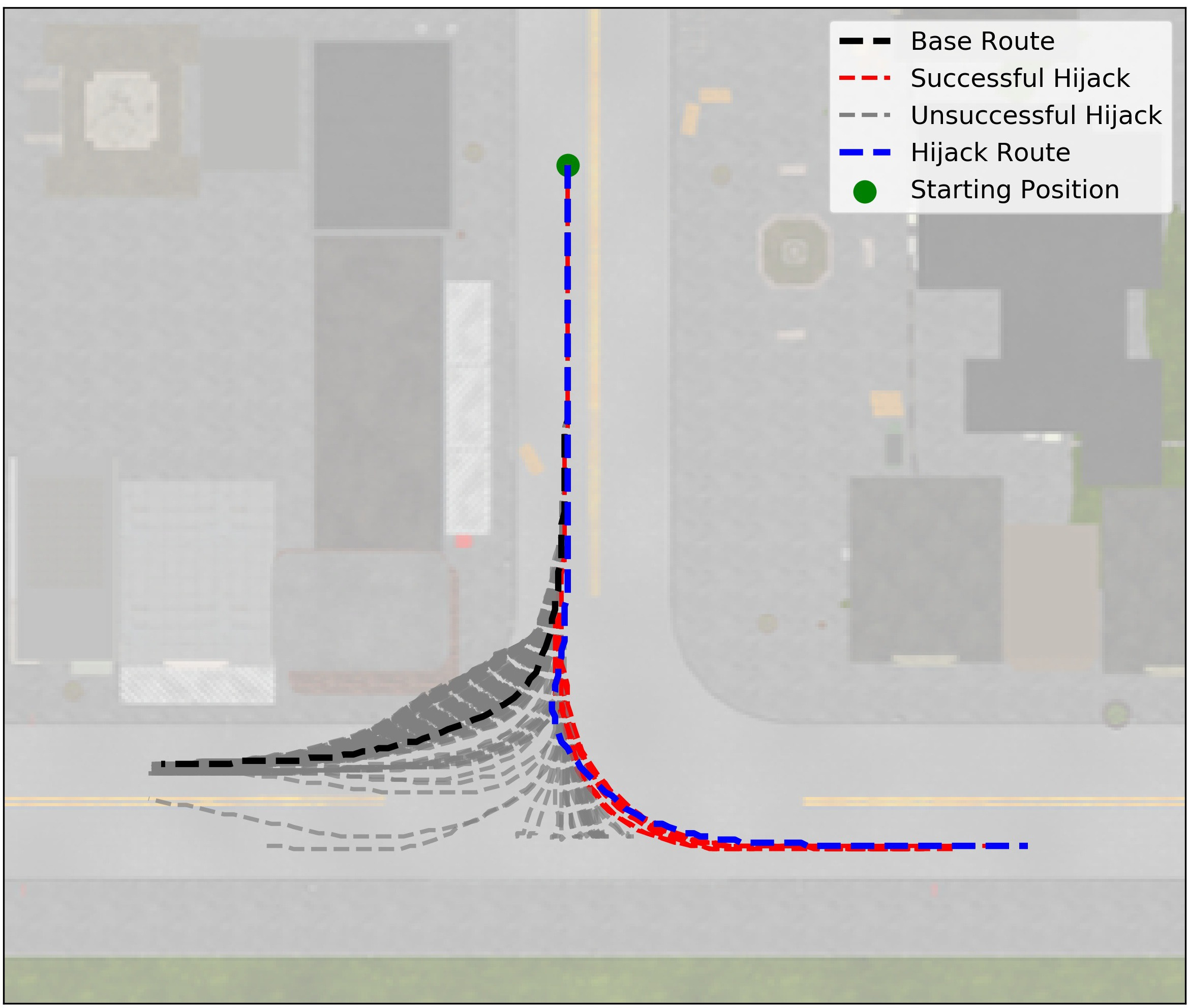}
  \caption{Illustration of a \textit{hijack attack} where we use an attack to trick the vehicle to deviate from its normal path (base route) to a target hijack route. It demonstrates a successful hijack where we make a vehicle otherwise turning right at an intersection, to turn left.}
  \label{fig:jsa_hijack_attack}
\end{figure}

\begin{table}[]
\centering
\caption{Success rate of Hijacking Attacks for six scenarios.}
\label{tab:hijack_success}
\begin{tabular}{lcc}
\hline
Hijack Success Rates         & \% Successful & \% Unsuccessful \\ \hline
Straight $\rightarrow$ Right & 14.8          & 85.2            \\
Straight $\rightarrow$ Left  & 0.0           & 100.0           \\
Left $\rightarrow$ Straight  & 23.7          & 76.3            \\
Left $\rightarrow$ Right     & 14.3          & 85.7            \\
Right $\rightarrow$ Left     & 1.4           & 98.6            \\
Right $\rightarrow$ Straight & 25.9          & 74.1            \\ \hline
\end{tabular}
\end{table}

Thus far, our exploration of adversarial examples against autonomous driving models focused on causing the car to crash, or cause other infractions. We now explore a different kind of attack: vehicle hijacking.
In this attack, the primary purpose is to stealthily lead the car along a target path of the adversary's choice.

When attacking the IL model, previous experiments have only targeted the \textit{Lane Follow} branch of this model. Now, we focus our attacks on three different branches of the IL Model: \textit{Right Intersection, Left Intersection,} and \textit{Straight Intersection}. Here, we define a successful attack to be an adversary that 1) causes no infractions or collisions and 2) causes the agent to make a turn chosen by the attacker rather than the ground truth at a particular intersection (e.g. the attacker creates an adversary to make the agent turn left instead of go straight through an intersection). With this definition, an attack that causes the agent to incur an infraction is not considered to be a successful attack. In order to produce such attacks, we modify our experimental setup. After choosing a particular intersection, we run the simulation with no attack to record the baseline steering angles over the course of the episode. The high-level command provided by CARLA directs the agent to take a particular action at that intersection (for example, go straight). We then modify the CARLA high-level command to the direction desired by the attacker (for example, take a right turn). After running the simulation, we store these target steering angles over the entire episode. Finally, we revert the CARLA high-level command to the original command provided to the agent during the baseline simulation run and begin generating attacks at the intersection. We modify our optimization problem to minimize the difference in the steering angles recorded during an episode with an attack ($\vec{\Theta}_{\delta}$ as defined in \ref{subsec:candidate_objective_functions}) and the steering angles of the target run (defined as $\vec{\Theta}_{\text{target}}$):
%This implies that our optimizer would generate adversaries that caused the agent to take the desired route.
\begin{subequations}
\label{E:attack4}
\begin{align}
\min_{l,\delta} ||\vec{\Theta}_{\delta} - \vec{\Theta}_{\text{target}}{||}_1 \label{E:attack4a}\\
\mathrm{subject\ to:}\quad l \in L, \quad \delta \in S. \label{E:attack4b}
\end{align}
\end{subequations}

CARLA (v0.8.2) did not include a four-way intersection in their provided maps, which constrain our experiments to a three-way junction as shown in Fig. \ref{fig:jsa_hijack_attack}. Of the six possible hijacking configurations, we were able to generate adversaries that successfully hijacked the car to take a desired route rather than the baseline route for five configurations. For example, Fig. \ref{fig:jsa_hijack_attack} shows the car being hijacked to take a right turn instead of going straight. While we were able to produce attacks that incurred an infraction in each scenario shown in Fig. \ref{fig:jsa_hijack_attack} (the gray paths), these episodes did not count as successful hijacks as the car did not take the target route. Table \ref{tab:hijack_success} shows the rate of successful attacks for the six available hijacking scenarios in CARLA v0.8.2. To conclude, we were able to modify our optimization problem and generate adversaries at intersections which caused the agent to take a hijacking route, rather than the intended route.

\subsection{Interpretation of Attacks using DeConvNet} 
\label{subsec:interpretation_of_attacks_using_deconvnet}

\begin{figure}
  \centering  \includegraphics[width=0.9\columnwidth]{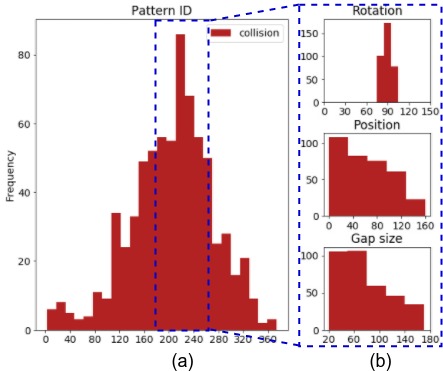}
  \caption{(a) Histogram showing strong adversaries. (b) Depiction of range of rotation, position and gap parameters for the most robust adversaries.}
  \label{fig:result2b}
\end{figure}

\begin{figure*}
  \centering  
  \includegraphics[width=\textwidth]{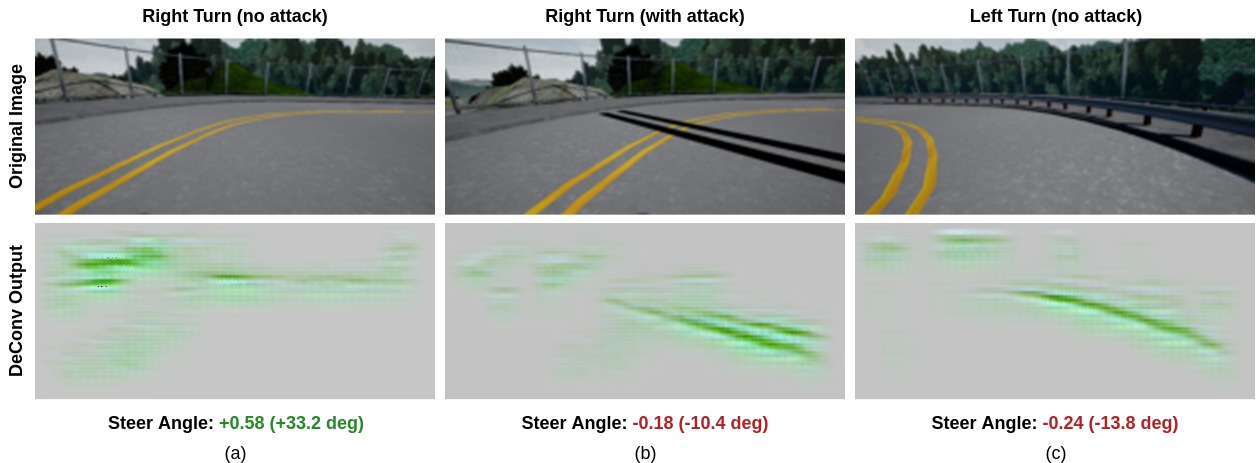}
  \caption{Attacks against Right Turn Driving: The top row shows the camera input while the bottom deconvolution images show that the reconstructed inputs from the strongest activations determine the steering angle. (a) Right Turn Driving without attack, (b) Right Turn Driving with attack and (c) Left Turn Driving without attack for comparison}
  \label{fig:result03}
\end{figure*}

%Through our previous experiments, we gain an intuition that some patterns perform better than others, 
In this section, our goal is to better understand what makes the attacks effective.
We begin by quantitatively analyzing the range of parameters of attacks that will generate the most robust attacks in the context of right turns.
%, i.e., attacks that would perform well against different environmental conditions for the  scenario of right turn driving. 
For simplicity, we analyze the Double Line attack whose parameters include rotation angle, position, and gap size.  Fig.~\ref{fig:result2b} shows a histogram of the collision incidence rates versus the pattern IDs, and its corresponding parameters for an experiment with 375 iterations. 
%We detect peaks in the histogram (Fig.~\ref{fig:result2b}(a)) which when dissected into constituent attack parameters as shown in 
Fig.~\ref{fig:result2b}(b), in particular, shows that some parameters play a stronger role than others in generating a successful attack. For example, in this particular setting Double Line attacks, successful adversaries have a narrow range of rotation angles (90 - 115 degrees). Fig.~\ref{fig:result2b}(b) also shows that smaller gap sizes perform slightly better than larger ones.  

%To develop a stronger, underlying intuition of why these attacks work, and why some of them work better than the others, we peel through the layers of the e2e imitation learning network.

To better understand the working mechanisms of the successful attack to the underlying imitation learning algorithm, we use network deconvolution, using 
%the activations of feature maps inside the network need to be interpreted. Interpreting the activations requires mapping the feature maps to the input layer, hence we adopt 
a state-of-the-art technique, DeConvNet \cite{Zeiler2014VisualizingAU}.
%to perform the mapping. 
Specifically, we attach each CONV block (a convolution layer with ReLU and a batch normalizer) to a DeConv counterpart, since the backbone of the imitation learning algorithm is a convolutional neural network which consists of eight CONV blocks for feature extraction and two fully connected (FC) blocks for regression. Each DeConv block uses the same filters, batch norm parameters, and activation functions as the CONV block, except that the operations are reversed. 
In this paper, DeConvNet is used merely as a probe to the already trained imitation learning network: it provides a continuous path to map high-level feature maps down to the input image. 
To interpret the network, the imitation learning network first processes the input image and computes the feature maps throughout the network layers. To view selected activations in the feature maps of a layer, other activations are set to zero, and the feature maps  backtrack through the rectification, reverse-batch norm, and transpose layers. Then, activations that contribute to the chosen activations in the lower layer are reconstructed. The process is repeated until the input pixel space is reached. Finally, the input pixels which give rise to the activations are visualized. In this experiment, we chose the \textit{top-200} strongest/largest activations in the \textit{fifth} convolution layer and mapped these activations down to the input pixel space for visualization. The reasons behind this choice are twofold: 1) The strongest activations stand out and dominate the decision-making in NNs and the \textit{top-200} activations are sufficient to cover the important activations, and 2) activations of the fifth CONV layer are more representative than other layers, since going deeper would mean that the amount of non-zero activations reduces significantly, which invalidates the deconvolution operations, while shallow layers fail to fully capture the relation between different extracted features. 

We conduct a case study to understand why an attack works. Specifically, we take a deeper look inside the imitation network when adversaries are attacking the autonomous driving model for the  right turn driving scenario. The baseline case without any attack is depicted in Fig.~\ref{fig:result03}(a) while the one with a successful double-line attack is shown in Fig.~\ref{fig:result03}(b). In the first row of Fig.\ref{fig:result03}, the input images from the front camera mounted on the vehicle are displayed, which are fed to the imitation learning network. In Fig. \ref{fig:result03}(a), the imitation learning network guides the vehicle to turn right at the corner, as the steering angle output is set to a positive value (steering +0.58). The highlighted green regions in the reconstructed inputs in the corresponding second row show the imitation network makes this steering decision mainly following the curve of the double yellow line. However, when deliberate attack patterns are painted on the road as shown in Fig. \ref{fig:result03}(b), the imitation network fails to perceive the painted lines which could be easily ignored by a human; instead, the network regards the lines as physical barriers and guides the vehicle to steer left (steering -0.18) to avoid a fictitious collision, leading to an actual collision. The reconstructed image below confirms that the most outstanding features are the painted adversaries instead of the central double yellow lines. We speculate that the vehicle recognizes the adversaries as the road curb. And Fig. \ref{fig:result03}(c) confirms our speculations. In this case, the vehicle is turning left and the corresponding reconstructed image shows the curb would contribute the strongest activations in the network which will make the steering angle a negative value (steering -0.24) to turn left. The similarity of the reconstructed inputs between cases (b) and (c) suggests that the painted attacks are misrecognized as a curb which leads to an unwise driving decision. To summarize, the deliberate adversaries that mimic important road features are very likely to be able to successfully attack the imitation learning algorithm. This also emphasizes the importance of taking more diverse training samples into consideration when designing autonomous driving techniques. Note that since the imitation learning network makes driving decisions solely based on current camera input, using one frame per case for visualization is enough to unravel the root causes of an attack's success.

\section{Conclusion}
\label{sec:conclusion}

In this paper, we develop a versatile modeling framework and simulation infrastructure to study adversarial examples on e2e autonomous driving models.
Our model and simulation framework can be applied beyond the scope of this paper, providing useful tools for future research to expose latent flaws in current models with the ultimate goal of improving them.
Through comprehensive experiment results, we demonstrate that simple physical adversarial examples that are easily realizable, such as mono-colored single-line and multi-line patterns, not only exist, but can be quite effective under certain driving scenarios, even for models that perform robustly without any attacks. 
We demonstrate that Bayesian Optimization coupled with a strong objective function is an effective approach to generating devastating adversarial examples. We also show that by modifying the objective function, we are able to hijack a vehicle where we cause the driverless car to deviate from its original route to a route chosen by an attacker. 
Finally, our analysis using the DeConvNet method offers critical insights to further explore attack generation and defense mechanisms. Our code repository is available at:

\href{https://github.com/xz-group/AdverseDrive}{\textbf{\textit{https://github.com/xz-group/AdverseDrive}}}.

\section{Acknowledgements}
We would like to thank Dr. Ayan Chakrabarti for his advice on matters related to computer vision with this research and Dr. Roman Garnett for his suggestions regarding Bayesian Optimization. We would also like to thank the CARLA team for their technical support regarding the CARLA simulator. This research was partially supported by NSF grants CNS-1739643, IIS-1905558 and CNS-1640624, ARO grant W911NF1610069 and MURI grant W911NF1810208.

% \section*{References}
\bibliographystyle{ieeetr}
\bibliography{bibliography}
\end{document}